\documentclass{ifacconf}

\makeatletter
\let\old@ssect\@ssect 
\makeatother

\usepackage{natbib}
\usepackage[colorlinks=true,allcolors=steelblue]{hyperref}

\makeatletter
\def\@ssect#1#2#3#4#5#6{%
  \NR@gettitle{#6}
  \old@ssect{#1}{#2}{#3}{#4}{#5}{#6}
}
\makeatother

\usepackage{graphicx}
\usepackage{amsmath}
\usepackage{optidef}

\usepackage{amssymb}
\usepackage{amsfonts} 
\usepackage{stmaryrd}
\usepackage{multicol}
\usepackage{cuted}
\usepackage{multirow}
\usepackage{caption, subcaption}
\newtheorem{theorem}{Theorem}

\newtheorem{definition}{Definition}

\usepackage{xcolor}
\usepackage{comment}
\usepackage{algorithm}
\usepackage{algorithmic}
\usepackage{scalerel}
\usepackage{bm}
\usepackage{url}
\usepackage{svg}

\definecolor{steelblue}{RGB}{70,130,180}

\usepackage[utf8]{inputenc}

\begin{document}

\begin{frontmatter}

\title{A Cantor-Kantorovich Metric Between Markov Decision Processes with Application to Transfer Learning\thanksref{footnoteinfo}} 

\thanks[footnoteinfo]{This project has received funding from the European Research Council (ERC) under the European Union’s Horizon 2020 research and innovation program under grant agreement No 834142 (Scalable Control), and grant agreement No 864017 (L2C). A. Banse is supported by the French Community of Belgium in the framework of a FNRS/FRIA grant, V. Renganathan is a member of the ELLIIT Strategic Research Area in Lund University and Raphaël M. Jungers is a FNRS honorary Research Associate.}

\author[First]{Adrien Banse} 
\author[Second]{Venkatraman Renganathan} 
\author[First]{Raphaël M. Jungers}

\address[First]{ICTEAM Institute, UCLouvain, Louvain-la-Neuve, 1348, Belgium. E-mail: \{adrien.banse, raphael.jungers\}@uclouvain.be.}
\address[Second]{Department of Automatic Control - LTH, Lund University, Sweden. E-mail: venkatraman.renganathan@control.lth.se}

\begin{abstract}     
We extend the notion of Cantor-Kantorovich distance between Markov chains introduced by \cite{Banse2023} in the context of Markov Decision Processes (MDPs). The proposed metric is well-defined and can be efficiently approximated given a finite horizon. Then, we provide numerical evidences that the latter metric can lead to interesting applications in the field of reinforcement learning. In particular, we show that it could be used for forecasting the performance of transfer learning algorithms.
\end{abstract}


\end{frontmatter}

\section{Introduction}
Research on quantitative notion of behavioural distance between Markov processes by the reinforcement learning community (see \citep{Garcia2022} and the references therein) mimics the study of distance between dynamical systems conducted by the control community (see \citep{el1985gap}, \citep{georgiou1988computation} and the references therein). Both communities are interested in computing \emph{how much processes/dynamical systems differ in terms of their behaviour}. Several metrics have been proposed for Markov Chains (MC) (see \citep{Kiefer2018}, \citep{Chen2014}, \citep{Rached2004}), including the recent Cantor-Kantorovich metric by \citet{Banse2023} where they applied it for abstraction-based methods. Few metrics like the one in \citep{Banse2023} are equipped with the availability of algorithms for fast computation making them deployment-ready. 

It is typical to train a reinforcement learning algorithm in a simpler world modeled as a Markov Decision Process (MDP) and deploy it in a real world setting corresponding to a different MDP. Several Transfer Learning (TL) algorithms have been developed in this paradigm where one transfers a learned policy from one MDP to another in the hope of improving the performance of the latter (see \citet{Lazaric2008}, \citet{Wang2020}, \citet{Tao2020}, \citet{inproceedings}). Many works have shown numerical evidences that TL algorithms have better performances when the source and target MDPs are similar to each other (see \citet{Song2016}, \citet{Carroll}, \citet{Zhu2023} and the references therein). This triggered the research community to undertake similar efforts as that of MC for studying the similarity between the MDPs. Interestingly \citet{Carroll} argued that no single task similarity measure is uniformly superior for TL problems. Despite their observation, it would be beneficial to design a metric between MDPs which does not have computational complexity issues so that it can be used to improve the performance of TL problems. 

This manuscript extends the range of application of the Cantor-Kantorovich metric proposed by \citet{Banse2023} in the context of TL. Specifically, the main contributions are: 
\begin{enumerate}
    \item The Cantor-Kantorovich metric is formulated in the context of MDPs.
    \item The promising potential of the proposed metric is demonstrated on a transfer learning problem where sources having smaller Cantor-Kantorovich distance with the target are shown to guarantee performance using TL techniques.
\end{enumerate}

\textbf{Outline:} Following a short summary of notations and preliminaries, we present the proposed metric between MDPs in Section \ref{sec_problem_formulation}. Subsequently, the application to problems in Transfer Learning domain is demonstrated in Section \ref{sec:TL}. Finally, the paper is summarised in Section \ref{sec_conclusion} along with the discussion of potential future research directions.

\textbf{Notations} The set of real and natural numbers are denoted by $\mathbb{R}$ and $\mathbb{N}$ respectively. For $N \in \mathbb{N}$, we denote by $[N] := \{0,1,\dots,N \}$. The cardinality of the set $C$ is denoted by $\left | C \right \vert$. The notation $f: X \rightarrow Y$ denotes that $f$ is a mapping from domain $X$ to range $Y$. Given a set $S \subseteq \mathbb{R}^{n}$, the Borel set associated with it is denoted by $\mathcal{B}(S)$. For a random variable $x$, its expected value is given by the notation $\mathbb{E}[x]$.
Let $(\Omega, D)$ be a discrete and finite metric space equipped with metric $D$. Given two probability distributions $\mathbb{P} : \Omega \to [0, 1]$ and $\mathbb{Q} : \Omega \to [0, 1]$, the Kantorovich distance between them is defined as 
\begin{equation}
    K_D(\mathbb{P}, \mathbb{Q}) = \min_{\pi \in \Pi(\mathbb{P}, \mathbb{Q})} \sum_{\omega_1, \omega_2 \in \Omega} 
    D(\omega_1, \omega_2)\pi(\omega_1, \omega_2), 
\end{equation}
where $\Pi(\mathbb{P}, \mathbb{Q})$ denotes the set of all joint probability distributions on $\Omega \times \Omega$ with $\mathbb{P}$ and $\mathbb{Q}$ being the marginals.

\section{The Cantor-Kantorovich metric in the context of MDPs}
\label{sec_problem_formulation}

In this section, we recall the notions introduced in \citep{Banse2023} and we show how to use them in the context of MDPs.


\subsection{Preliminaries About Markov Decision Processes}

The following formalism in inspired from \citep{Abate2013}.

\begin{definition}[MDP]
A Markov Decision Process is described using a tuple $M = (\mathcal{S}, \mathcal{U}, \mathcal{T}, R, \mu)$, where $\mathcal{S}$ is the set of \emph{states}, $\mathcal{U}$ is the set of \emph{control actions}, $\mathcal{T}$ is the conditional stochastic kernel that assigns to each state $s \in \mathcal{S}$ and control action $u \in \mathcal{U}$, a probability measure $\mathcal{T}(\cdot \, | \, s, u) : \mathcal{B}(\mathcal{S}) \to [0, 1]$. Further, $R : \mathcal{S} \to \mathbb{R}$ denotes the reward function and $\mu : \mathcal{B}(\mathcal{S}) \to [0, 1]$ the initial measure on the states. 
\end{definition}
In this work, we consider that $\mathcal{U}$ is a finite set. For each $u \in \cal U$, we define $\tau_u(s, \cdot) := \mathcal{T}(\cdot \, | \, s, u)$ for all $s \in \mathcal{S}$. The above definition of MDP leads to the following semantics for a process $X(k)$ over the horizon $k \in [0, N]$ with $N \in \mathbb{N}$. Given a control action sequence $\mathbf{u}^N = (u_k)_{k \in [N-1]} \in \mathcal{U}^N$, and an initial state $s_0 \in \mathcal{S}$ sampled according to $\mu$, the semantics of the process $X(k)$ for $k \in [N-1]$ is defined as
\begin{align} \label{eqn_process}
X(k + 1) \sim \mathcal{T}(\cdot \, | \, X(k), u_k) 
\quad 
\text{with}
\quad 
X(0) = s_0.
\end{align}

Given a MDP $M = (\mathcal{S}, \mathcal{U}, \mathcal{T}, R, \mu)$, its trajectory of length $N$ is denoted by $\mathbf{s}^N := (s_k)_{k \in [N-1]}$. In this work, we consider that the control actions associated to this trajectory are determined by a policy $p : \mathcal{S} \to \mathcal{A}$. Given a MDP $M$, a policy $p$ and a horizon $N$, the probability distribution induced by the MDP $M$ at time $N$ is defined as 
\begin{equation} \label{eq:probability_induced}
    \mathbb{P}^N_{p}(\mathbf{s}^N) =  \mu(s_0) \prod_{i = 0}^{N-2} \tau_{p(s_i)}(s_i, s_{i+1})
\end{equation}
for every sequence $\mathbf{s}^N \in \mathcal{S}^N$. For a given policy $p$, the function $\mathbb{P}^N_p$ is therefore a discrete and finite probability distribution of dimension $|\mathcal{S}|^N$.


\subsection{A metric between dynamics of two MDPs}

We now recall the main recent results in \citep{Banse2023} in the context of MDPs. Consider two MDPs $M_1 = (\mathcal{S}_1, \mathcal{U}_1, \mathcal{T}_1, R_1, \mu_1)$ and $M_2 = (\mathcal{S}_2, \mathcal{U}_2, \mathcal{T}_2, R_2, \mu_2)$. We assume that both MDPs are homogeneous (see Definition 3.2 in \citep{Song2016}), meaning that there exist one-to-one correspondence between their state-spaces $(\mathcal{S}_1, \mathcal{S}_2)$, and also between their control spaces $(\mathcal{U}_1, \mathcal{U}_2)$. However, for the ease of exposition, we will proceed ahead with a simpler setting where $\mathcal{S}_1 = \mathcal{S}_2 = \mathcal{S}$ and $\mathcal{U}_1 = \mathcal{U}_2 = \mathcal{U}$. 

We are interested in the asymptotic difference between the dynamics of both MDPs under two policies $p$ and $q$. Given a horizon length $N$, consider the metric space $(\mathcal{S}^N, C)$, equipped with the Cantor metric $C$. Given two trajectory sequences $\mathbf{a}^N, \mathbf{b}^N \in \mathcal{S}^N$, the Cantor metric between $\mathbf{a}^N$ and $\mathbf{b}^N$ is defined as 
\begin{align}
C(\mathbf{a}^N, \mathbf{b}^N) 
:= 
2^{-\inf\{k : a_k \neq b_k\}}.    
\end{align}

Given a horizon $N$ and two policies $p$ and $q$, let $\mathbb{P}^N_{p}$ and $\mathbb{Q}^N_q$ be the two probability distributions respectively induced by MDPs $M_1$ and $M_2$ at time $N$ according to \eqref{eq:probability_induced}. The next theorem is an extension of \citep[Theorem~1]{Banse2023}, and shows that $K_C\left(\mathbb{P}^N_p, \mathbb{Q}^N_q\right)$ can be computed iteratively.
\begin{theorem} \label{thm:recursive}
    Given a horizon $N > 1$ and two policies $p$ and $q$, it holds that 
    \begin{equation} \label{eq:ck_rec}
    \begin{aligned}
        &K_C\left(\mathbb{P}^{N+1}_p, \mathbb{Q}^{N+1}_q\right) \\
        &= 
        K_C\left(\mathbb{P}^N_p, \mathbb{Q}^N_p\right)  \\
        &+ 
        2^{-(N+1)}
        \sum_{\mathbf{s}^N \in \mathcal{S}^N} 
        \left( 
            r_{p, q}(\mathbf{s}^N) - \sum_{s \in \mathcal{S}} r_{p, q}(\mathbf{s}^N s)
        \right), 
    \end{aligned}
    \end{equation}
    with $r_{p, q}(\mathbf{s}^N) = \min\left\{ \mathbb{P}^N_p(\mathbf{s}^N), \mathbb{Q}^N_q(\mathbf{s}^N)\right\}$.
\end{theorem}

Given two policies $p$ and $q$, we define the metric the dynamics of two MDPs as 
\begin{equation} \label{eq:dist_dynamics}
    \mathbf{d}(M_1, M_2) := \lim_{N \to \infty} K_C\left(\mathbb{P}^N_p, \mathbb{Q}^N_q\right).
\end{equation}
It can be shown that this metric satisfies
\begin{equation}
    0 \leq \mathbf{d}(M_1, M_2) - K_C\left(\mathbb{P}^N_p, \mathbb{Q}^N_q\right) \leq 2^{-N}, 
\end{equation}
which allows the user to approximate this metric with a finite horizon $N$ (see \citep[Theorem~2]{Banse2023} for a similar result in the context of Markov Chains). Moreover, this metric can be efficiently approximated using a very similar algorithm as \citep[Algorithm~1]{Banse2023}.


To give the reader some intuition, the Cantor-Kantorovich metric $\mathbf{d}$ captures a \textit{discounted difference between the dynamics of the two MDPs} because the Cantor distance can be interpreted as a discount factor.

The choice of the specific policies $p$ and $q$ depends on the application and the specific example. In Section~\ref{sec:TL}, we propose a choice of policies for forecasting transfer learning performance. The investigation of more general choices is left for further work.

\section{Application to Transfer Learning} \label{sec:TL}
\begin{figure}[ht!]
    \centering
    \includegraphics[width = 0.6\linewidth]{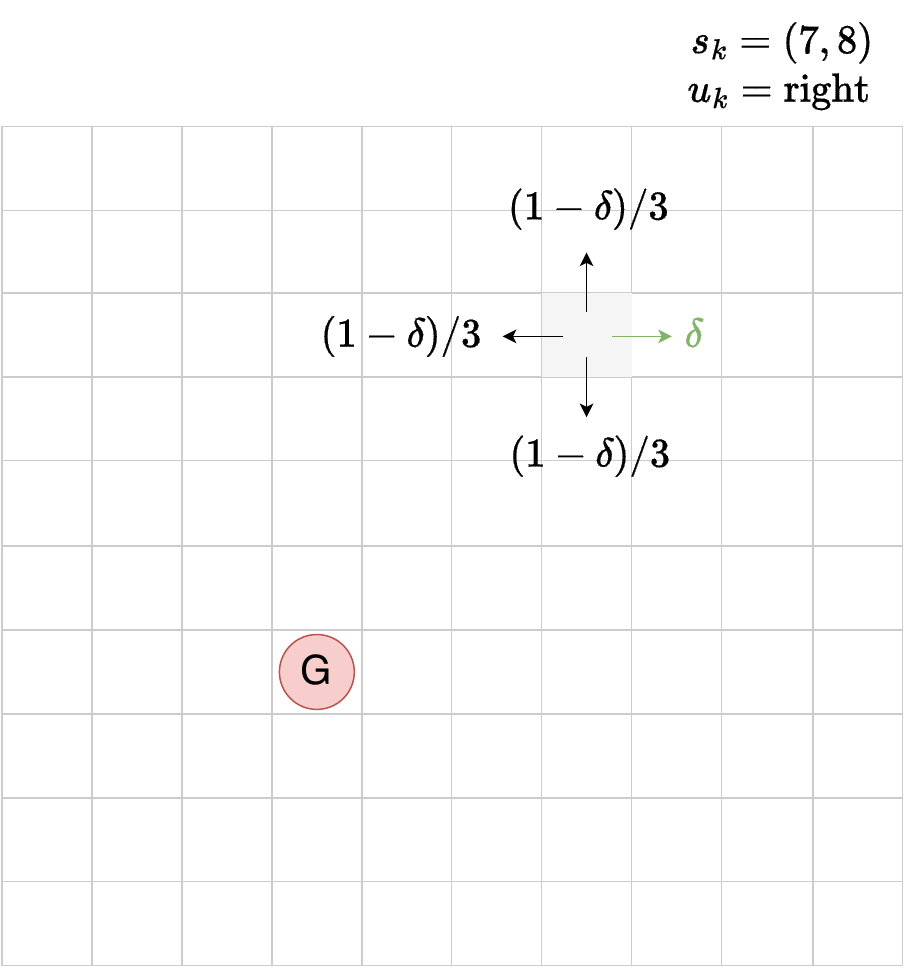}
    \caption{A grid-world of size $10 \times 10$ with a goal in $(4, 4)$. If a time $k$, the action $u_k = \text{right}$ is chosen, then the probability of going in this direction is $\delta$, and the probability to go in other directions is $(1-\delta)/3$.}
    \label{fig:gridworld}
\end{figure}

In this section, we showcase the possible application of our Cantor-Kantorovich distance to transfer learning with a numerical experiment. We consider a grid-world target MDP which is illustrated in Figure~\ref{fig:gridworld}. The size of the grid-world is $10 \times 10$; the possible control actions are $ \mathcal{U} = \{\text{left}, \text{right}, \text{up}, \text{down}\}$; the goal is in position $(4, 4)$, and corresponds to a reward of 10; when choosing a direction $u$, the probability to go in that direction is $\delta = 1/2$, and the other probabilities are $(1-\delta)/3 = 1/6$.

The transfer learning experiment goes as follows. We generate 100 other grid-worlds $M_{S, i}$, where the only difference with the target is the probability $\delta$, uniformly sampled between 0 and 1. For each source MDP $M_{S, i}$:
\begin{enumerate}
    \item We compute an optimal policy $p^*_{S, i}$ with Q-learning and we save the optimal Q-table $Q^*_{S, i}$. Specifically, we used an $\varepsilon$-greedy exploration strategy with $\varepsilon = 0.5$, and we learned the optimal policy with 4000 episodes of length 100, and with a learning rate of 0.01. We approximated the rewards with 10000 samples, and the discount factor is $\gamma = 0.95$.
    \item We compute the Cantor-Kantorovich distance with the target, that is $\mathbf{d}(M_T, M_{S, i})$ using $N = 8$. We fix the policies $p$ and $q$ as 
    \begin{equation}
        p = q = p^*_{S, i}.
    \end{equation}
    \item We solve the target with Q-learning by initializing Q-values with $Q^*_{S, i}$, and we compute the jump-start reward (which describes the increase in the initial performance achievable in the target using the transferred knowledge, before any further learning), that is the difference of reward at the start of the learning process with and without transfer learning.
\end{enumerate}

\begin{figure}[ht!]
    \centering
    \includesvg[width = \linewidth]{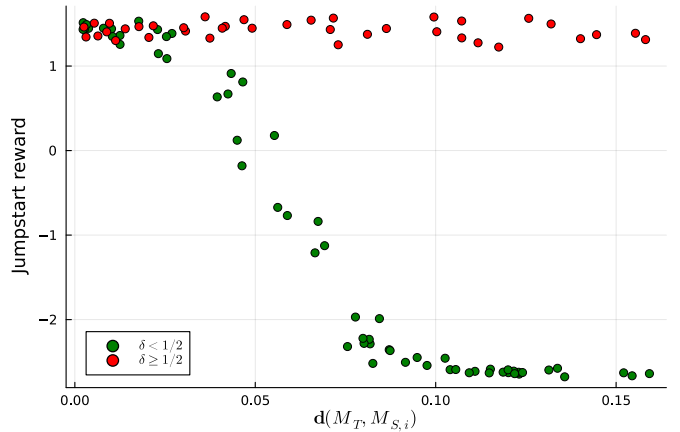}
    \caption{Results of the transfer learning experiment. The x-axis is the Cantor-Kantorovich distance between the target and the sources. The y-axis is the jumpstart metric, i.e. the metric used to asses the performance of the transfer. Green and red dots correspond to sources with $\delta < 1/2$, and $\delta \geq 1/2$ respectively.}
    \label{fig:TL}
\end{figure}

The results of this experiment can be found in Figure~\ref{fig:TL} and they can be reproduced with the code provided in \url{https://github.com/adrienbanse/MTNSExperiments}.

First we can observe that, in the red case (i.e. when the source has a greater chance to take the right direction than the target), the TL technique used here always improves performance as the jumpstart reward is always larger than 1. This is due to the fact that the policy found by the source is always optimal for the target as well. In the second case though, the green dots show that there is a strong correlation between the Cantor-Kantorovich distance and the performance of the transfer. 

This experiment therefore provides a numerical evidence that sources with a non-optimal policy but with a small Cantor-Kantorovich distance with the target guarantee performance using transfer learning. Although the example discussed here is simple, it is rich enough to demonstrate the effectiveness of our proposed metric. We leave the option of exploring a more involved numerical example with different reward setting, studying other TL algorithms and other nuances to the future study.

\section{Conclusion \& Future Outlook} \label{sec_conclusion}

In this work, a novel Cantor-Kantorovich metric with reasonable computational complexity was introduced in the context of MDPs. Its applicability to problems in the TL domain was also demonstrated using a simple numerical simulation.  

There are several promising and potential research directions for the future. For instance, one could aim for improving the upper bound for accuracy of the proposed Cantor-Kantorovich metric with a finite horizon $N$. Similarly, one could investigate a distance of the form 
\begin{equation}
    \mathbf{d}'(M_1, M_2) = \alpha \mathbf{d}(M_1, M_2) + \beta \mathbf{d}_r(M_1, M_2), 
\end{equation}
where $\mathbf{d}_r$ is a distance between the rewards of $M_1$ and $M_2$. The choice of $\mathbf{d}_r, \alpha$ and $\beta$ would therefore depend on the application context. For example, one could investigate the distances described in \citep{Gleave2020}. In the same fashion as \cite{Carroll}, one could apply such distances to the same grid-world example as above, but where the goal is moving. Finally, it would be interesting to investigate other performance measures than the jumpstart reward to evaluate the performance of the proposed metric in the transfer learning setting.





\bibliography{main}

\end{document}